\pdfoutput=1

\documentclass[11pt]{article}
\usepackage[]{acl}

\usepackage{times}
\usepackage{latexsym}

\usepackage[T1]{fontenc}

\usepackage[utf8]{inputenc}

\usepackage{microtype}
\usepackage{enumitem}
\usepackage{amssymb}
\usepackage{amsmath}
\usepackage{makecell}
\usepackage[caption=false]{subfig}

\usepackage{graphicx}
\usepackage{algorithm}
\usepackage{algorithmicx}
\usepackage{algpseudocode}
\usepackage{tablefootnote}
\usepackage{verbatim}

\usepackage{multirow}

\title{Multi-Type Conversational Question-Answer Generation \\with Closed-ended and Unanswerable Questions}

 \author{Seonjeong Hwang$^{1}$, Yunsu Kim$^{1,2}$, Gary Geunbae Lee$^{1,2,}$ \\
         $^{1}$ Graduate School of Artificial Intelligence, POSTECH, Pohang, South Korea \\ $^{2}$ Computer Science and Engineering, POSTECH, Pohang, South Korea \\
         \texttt{\{seonjeongh, yunsu.kim, gblee\}@postech.ac.kr}}

\begin{document}

\maketitle
\begin{abstract}

Conversational question answering (CQA) facilitates an incremental and interactive understanding of a given context, but building a CQA system is difficult for many domains due to the problem of data scarcity.
In this paper, we introduce a novel method to synthesize data for CQA with various question types, including open-ended, closed-ended, and unanswerable questions.
We design a different generation flow for each question type and effectively combine them in a single, shared framework.
Moreover, we devise a hierarchical answerability classification (hierarchical AC) module that improves quality of the synthetic data while acquiring unanswerable questions.
Manual inspections show that synthetic data generated with our framework have characteristics very similar to those of human-generated conversations.
Across four domains, CQA systems trained on our synthetic data indeed show good performance close to the systems trained on human-annotated data.
\end{abstract}

\section{Introduction}

Conversational question answering (CQA) aims to answer a question based on a given passage and previous conversation. Unlike single-turn question answering (QA) \cite{rajpurkar2016squad,rajpurkar2018know,kwiatkowski2019natural}, CQA encourages questioners to incrementally make follow-up questions, which is suitable for services that require active interaction between humans and systems. However, manually creating large amounts of conversations is very costly, which is a barrier to its utilization in various domains.

To alleviate this issue, a few methods for conversational question generation have been studied \cite{gao2019interconnected,pan2019reinforced,nakanishi2019towards,shen2021gtm,gu2021chaincqg}. Furthermore, we have proposed approaches for automatically synthesizing multi-turn conversational question-answer (Q--A) pairs in order to build training data for CQA in our previous studies \cite{hwang2021study,hwang2022conversational}. However, our previous frameworks generate only open-ended questions that cannot be answered succinctly. In real-world situations, concise answers, such as \emph{yes}, \emph{no}, and \emph{unknown}, are essential for fast interaction and simplified conversations.

In this paper, we introduce MultiCQAG, a framework that can generate multiple types of CQA data. To enable this, we insert a generation flow for closed-ended Q--A pairs to our previous framework \cite{hwang2022conversational}. We also design a hierarchical answerability classification (hierarchical AC) module that collects yet another type of data --- unanswerable questions --- while improving data quality by removing invalid Q--A pairs.

In experiments, CQA systems trained on our synthetic datasets achieve an average F1 score of 77.2\% for four new domains, showing a difference of only 5.4\% from those trained on human-annotated data. Moreover, we show by manual evaluation that our synthetic data have a data distribution similar to that of human-annotated data. 

The contributions of this work can be summarized as follows:
\begin{itemize}
  \item We propose MultiCQAG, which synthesizes a CQA data consisting of various types of questions, including open-ended, closed-ended, and unanswerable questions.
  \item We design a hierarchical AC algorithm that filters out invalid Q--A pairs and acquires unanswerable questions.
\end{itemize}

\section{\label{tab:2}Background}

\begin{figure*}[h!]
\centering
\includegraphics[width=\textwidth]{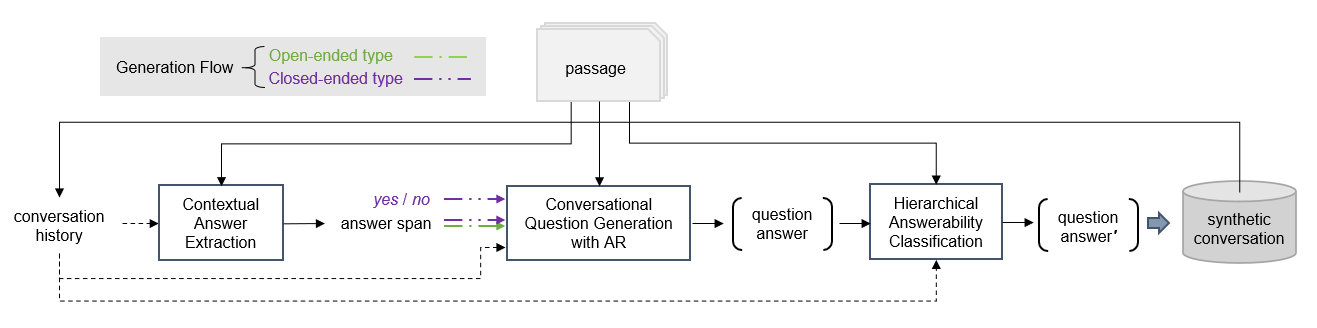}
\caption{\label{tab:Main achitecture} Generation pipeline of MultiCQAG. Conversation history is not used to generate the first Q--A pair of a conversation (dotted line).}
\end{figure*}

In our previous study, we proposed a conversational question-answer generation (CQAG) framework that automatically synthesized data for CQA given passages and that consisted of two modules: contextual answer extraction (CAE) and conversational question generation (CQG) \cite{hwang2021study}. First, the CAE module extracts a potential answer span from a passage based on a previous conversation. Second, the CQG module generates a conversational question for the extracted answer. During generation, the framework uses previously generated Q--A pairs as the conversation history for the next generation. However, synthetic data generated by this framework only contain extractive answers that are inflexible in form. Moreover, there is a risk that errors generated by the CAE module can propagate to subsequent generations.

To resolve this problem, we developed CQAG-AR, which adopted an answer revision approach \cite{hwang2022conversational}. In this framework, the CQG with answer revision (CQG-AR) module generates a question for the extracted answer span and then modifies the answer span so that it better fits the question. However, CQAG-AR can only synthesize open-ended types of data and cannot generate closed-ended and unanswerable types, which are frequently used in human conversations. In this paper, we improve CQAG-AR to generate those different types of data in a single framework.

\section{Method}
\subsection{\label{tab:3.1} Generation Flows}
As shown in Figure \ref{tab:Main achitecture}, we insert two generation flows between CAE and CQG-AR modules to generate open-ended and closed-ended data. The CAE module $P(a^s|p,h;\theta_A)$ extracts an answer span $a^s$ that is a question worthy phrase in the passage $p$ considering the conversation history $h$, which is the concatenation of previously generated Q--A pairs. After extracting the answer span, the data type to generate for the current turn is randomly selected according to a preset ratio (open-ended:\emph{yes}:\emph{no}).

When the open-ended type is selected, the CQG-AR module generates an open-ended question $q^{open}$ and a revised answer $a^r$ for the answer span $a^s$ with consideration for the answer context $c^a$ and conversation history $h$, i.e., $P(q^{open}, a^r \mid c^a, h, a^s ; \theta_Q)$, where the answer context indicates the chunk of the passage containing the answer span and $N$ words front of and behind it. When the closed-ended type is chosen, however, the module generates a closed-ended question $q^{close}$ for \emph{yes} or \emph{no} based on the answer context and conversation history, i.e., $P(q^{close} \mid c^a, h, \mathrm{\emph{yes}}$/$\mathrm{\emph{no}} ; \theta_Q)$.

We implement both modules the same as in CQAG-AR.
However, in MultiCQAG, the two generation flows share the same model parameters $\theta_Q$ of the CQG-AR module, and the answer revision is only conducted for open-ended data.
Therefore, the module is trained to return the same answer (\emph{yes}/\emph{no}) as the input instead of a revised answer for closed-ended data.

\subsection{\label{tab:3.2} Hierarchical Answerability Classification}

Our framework has an autoregressive pipeline over multiple turns, so if an inappropriate Q--A pair is synthesized, the errors can propagate to subsequent data generation. Therefore, we devise a hierarchical AC module that determines whether a question can be answered based on the passage. If not, the module replaces the answer of an unanswerable question with "unknown".

\subsubsection{Algorithm}

\begin{algorithm}
\renewcommand{\algorithmicrequire}{\textbf{Input:}}
\renewcommand{\algorithmicensure}{\textbf{Output:}}
\caption{Hierarchical Answerability Classification}\label{tab:algo}
\begin{algorithmic}[1]
\Require question $q$, answer $a$, passage $p$, threshold $\tau$, classifier $f$
\Ensure ($q$, $a$) or ($q$, unknown) or DISCARD
\State tokenize $p$ into sentences $S$ = \{$s_1$, $s_2$, ... , $s_{|S|}$\}
\State get context sentence $c \in S$
\If{$f(q, c) > \tau$} \Comment{Context-level classification}
        \State \textbf{return} ($q$, $a$) 
\Else  \Comment{Passage-level classification}
        \State $L_{prob} \leftarrow \{f(q,s) | s \in S \setminus \{c\}\}$
        \If{$\max(L_{prob}) > \tau$}
                \State \textbf{return} DISCARD 
        \Else
                \State \textbf{return} ($q$, unknown)
        \EndIf
\EndIf
\end{algorithmic}
\end{algorithm}

We classify synthetic questions into three categories: (1) \emph{answerable in correct context} or an answerable question given the context sentence of the synthetic answer, (2) \emph{answerable in different context} or a question whose correct answer can be found in a sentence outside the context of its synthetic answer, and (3) \emph{unanswerable} question or a question that cannot be answered with the information in the passage.

Algorithm \ref{tab:algo} shows the hierarchical AC. The classifier $f$ returns the probability that a given context sentence answers the question. If the probability is over the threshold $\tau$, the question is considered an answerable question. 

In \textbf{context-level classification}, questions belonging to the \emph{answerable in correct context} category are detected. If the question belongs to this class, we keep the Q--A pair as it is. Otherwise, we proceed with \textbf{passage-level classification}, where the question is compared with all sentences in the passage except for $c$. If any of them contain the correct answer to the question, it means that the question is paired with the wrong answer (\emph{answerable in different context}). Thus, we discard this kind of Q--A pair. Questions other than these two types are \emph{unanswerable} question, and their answers are replaced with "unknown".

\subsubsection{\label{tab:3.2.2}Modeling}

The classifier $f$ is a model for natural language inference (NLI) tasks. Specifically, we implement it using ALBERT \cite{lan2019albert}, the current state-of-the-art model for question-answering NLI, which is a task to determine whether a text answers a question. We use QNLI \cite{wang2018glue}, a dataset for question-answering NLI, and CoQA \cite{reddy2019coqa}, a dataset for CQA, to train the classifier. Since the proportion of unanswerable questions in CoQA is only 1.3\%, we pre-train the model with QNLI and then fine-tune it with CoQA. 

To fully understand a conversational question, it is necessary to consider the previous conversation as well. Thus, we configure the classifier's input as $h$;\textit{<Q>};$q$;$s$ where $s$ indicates the sentences in the passage. The conversation history $h$ is omitted during pre-training because QNLI is based on single-turn QA. In addition, we insert a special token \textit{<Q>} in front of the question to distinguish it from the questions in the conversation history. To alleviate data imbalance in CoQA, we add negative samples by paring every sentence in the passage with an unanswerable question and train the model to minimize the focal loss \cite{lin2017focal} to learn more intensively on misclassified samples.

\section{Experiments}

We utilized CoQA \cite{reddy2019coqa}, a dataset for constructing CQA systems, to prove that MultiCQAG generates high-quality synthetic CQA data in new domains. CoQA is based on passages collected from seven different domains. Among the domains, only five that constituted official training and development sets\footnote{https://stanfordnlp.github.io/coqa/} were used in our experiments: Wikipedia, for training MultiCQAG, and four other domains for data synthesis and CQA evaluation.

\begin{table}[h!]
\centering
\scriptsize
\renewcommand{\arraystretch}{1.2}
\renewcommand{\tabcolsep}{4pt}
\begin{tabular}{lll|rl}
\hline
\multicolumn{3}{l|}{Data type}                                                                                                                                  & \multicolumn{1}{r|}{\#Q--As (Percentage)} & Total                   \\ \hline
\multicolumn{1}{l|}{\multirow{3}{*}{Answerable}} & \multicolumn{2}{l|}{Open-ended}                                                                     & \multicolumn{1}{r|}{20,354 (82.0\%)}     & \multirow{3}{*}{24,521} \\ \cline{2-4}
\multicolumn{1}{l|}{}                            & \multicolumn{1}{l|}{\multirow{2}{*}{\begin{tabular}[c]{@{}l@{}}Closed\\ -ended\end{tabular}}} & Yes & \multicolumn{1}{r|}{2,617 (10.5\%)}      &                         \\ \cline{3-4}
\multicolumn{1}{l|}{}                            & \multicolumn{1}{l|}{}                                                                         & No  & \multicolumn{1}{r|}{1,550 (6.2\%)}       &                         \\ \hline
\multicolumn{3}{l|}{Unanswerable}                                                                                                                      & \multicolumn{2}{r}{286 (1.2\%) $\rightarrow$ 2,957}                \\ \hline
\end{tabular}%
\caption{\label{tab:Data Distribution} Number of CoQA data examples in the Wikipedia domain used to train the modules of MultiCQAG.}
\end{table}

Table \ref{tab:Data Distribution} shows the statistics of CoQA data examples used to train the MultiCQAG modules. The CAE module was trained with 20,354 open-ended examples, and the CQG-AR module learned 24,521 answerable examples. When the AC model was trained, we used 24,521 answerable examples and 2,957 unanswerable examples, which were augmented from the original 286 examples by applying the negative sampling method described in Section \ref{tab:3.2.2}.

We generated synthetic CQA data for four domains (children's stories, literature, news, and middle and high school English exams) by using trained MultiCQAG and four collections of passages extracted from CoQA training and development sets. See Appendix \ref{tab:data examples} for examples of synthetic data and Appendix \ref{tab:hyperparameters} for implementation details.

The synthetic data were evaluated through two methods: extrinsic evaluation using the CQA task, and human evaluation. During the extrinsic evaluation, we trained T5-based CQA systems for four different domains on our synthetic data, and we evaluated each system using the CoQA test set\footnote{https://github.com/google/BIG-bench} for corresponding domains.

\subsection{\label{tab:ablation study section} Main Results}

To investigate the impact of our methods on CQA data generation, we sequentially added each component to CQAG-AR \cite{hwang2022conversational}, which is described in Section \ref{tab:2}. We generated synthetic data using these frameworks, and then trained CQA systems with each generated data. Table \ref{tab:Main Results} reports F1 scores of the CQA systems on the CoQA test set.

\begin{table}[h!]
\centering
\scriptsize
\renewcommand{\arraystretch}{1.2}
\renewcommand{\tabcolsep}{4pt}
\begin{tabular}{llcccc}
\hline
\multirow{2}{*}{Data}      & \multirow{2}{*}{Framework}         & \multicolumn{4}{c}{Domain}      \\ \cline{3-6} 
                           &                                    & Child. & Liter. & News. & Exam. \\ \hline
\multirow{4}{*}{Synthetic} & CQAG-AR (baseline)                        & 57.1   & 57.5   & 68.7  & 62.0  \\
                           & + Closed-ended generation & \underline{76.1}   & \underline{73.0}   & \textbf{81.5}  & \underline{77.3}  \\
                           & \,\,\,\,\, + Context-level AC                 & 73.9   & 71.6   & 80.2  & 75.6  \\
                           & \,\,\,\,\,\,\,\,\,\,\,\,+ Passage-level AC     & \textbf{76.7}   & \textbf{73.3}   & \underline{81.3}  & \textbf{77.5}  \\ \hline
Real            & CoQA                            & 83.8   & 80.2   & 85.2  & 81.1  \\ \hline
\end{tabular}%
\caption{\label{tab:Main Results} F1 scores on the CoQA test sets. Among the synthetic data results, the best results are in bold, and the second-best results are underlined.}
\end{table}

Adding the closed-ended data generation flow leads to F1 score improvements of about 15.7\% on average across all domains. These results demonstrate that our method is effective in generating data for training CQA systems that can answer closed-ended questions. Next, we applied each classification level of hierarchical AC. Performing only context-level AC degraded the performance of the CQA systems, but it was recovered by applying passage-level AC. This means that the answer to a question, which was considered unanswerable in the context-level AC despite it having a correct answer in another context, was replaced with "unknown" and confused the CQA system. 

Finally, when all our methods are combined, we achieve an F1 score of 77.2\% on average across the four domains. This improves the baseline result greatly by about 15.9\% and is only 5.4\% behind the system trained on costly human annotations.

\subsection{Analysis by Data Type}

To examine the contribution of our methods, we evaluate the CQA systems on each data type. Because the number of unanswerable data in the CoQA test set was quite small, we also used the development set for evaluation. In CoQA, a question has multiple answer candidates; we split the data based on the most frequent types of candidates.

\begin{table}[h!]
\centering
\scriptsize
\renewcommand{\arraystretch}{1.2}
\renewcommand{\tabcolsep}{4pt}
\begin{tabular}{llccc}
\hline
Data                       & Framework                & Open & Close & Unanswerable \\ \hline
\multirow{3}{*}{Synthetic} & CQAG-AR (baseline)                & 62.0 & 4.2   & 0.0     \\
                           & + Closed-ended generation & 61.3 & 74.6  & 0.0     \\
                           & \,\,\,\,\,+ Hierarchical AC        & 61.7 & 72.2  & 13.3     \\ \hline
Real                       & CoQA                     & 65.0 & 79.8  & 0.0     \\ \hline
\end{tabular}%
\caption{\label{tab:case study} Performance of the CQA systems by data type.}
\end{table}

As can be seen from the results in Table \ref{tab:case study}, the F1 score for closed-ended data is significantly better when the closed-ended generation flow is added to the baseline. Although the performance for open-ended types is slightly lower than before inserting the closed-ended generation flow, the change is insignificant. The result shows that both generation flows operate effectively within a single framework. 

With the addition of the hierarchical AC module, the system finally starts to respond to unanswerable questions. Note that even real training data from CoQA is insufficient to teach the model to handle unanswerable questions correctly (last row). Our synthetic data secures the unanswerable questions explicitly. We believe that the performance on unanswerable questions will improve further if we intentionally generate more questions with an "unknown" answer.

The minor degradation on closed-ended questions by hierarchical AC can be attributed to the small portion of closed-ended questions used for training the classifier (19.8\% of CoQA training set was comprised of closed-ended questions). We plan to remedy this by adjusting the data balance for each type in future work.

\subsection{Human Evaluation}

\begin{table*}[h]
\centering
\scriptsize
\renewcommand{\arraystretch}{1.2}
\begin{tabular}{ll}
\hline
\multicolumn{2}{l}{\textbf{Conversational Connectivity}: Whether questions are naturally connected to previous conversations.} \\ \hline
\multicolumn{1}{l|}{Dependent}          & Questions cannot be answered without previous conversations.              \\
\multicolumn{1}{l|}{Independent}        & Questions can be answered without previous conversations.                  \\
\multicolumn{1}{l|}{Unnatural}          & Questions have grammatical errors or overlaps with previous conversations.  \\ \hline
\multicolumn{2}{l}{\textbf{Question Answerability}: Whether answers can be found in a given passage.}                          \\ \hline
\multicolumn{1}{l|}{Answerable}         & Questions can be answered based on a given passage.                        \\
\multicolumn{1}{l|}{Unanswerable}       & Questions cannot be answered based on a given passage.                     \\ \hline
\multicolumn{2}{l}{\textbf{Answer Correctness}: Whether answers match the paired question.}                  \\ \hline
\multicolumn{1}{l|}{Correct}            & Questions are paired with correct answers.                            \\
\multicolumn{1}{l|}{Partially correct}  & Answers are incomplete or contain unnecessary information.              \\
\multicolumn{1}{l|}{Incorrect}          & Not the correct answer to the question.                                 \\ \hline
\end{tabular}%
\caption{\label{tab:assessment items} Assessment items for the human evaluation of CQA data. When the question is judged to be unnatural, the evaluation of other items is skipped.}
\end{table*}

\begin{table}[h!]
\centering
\scriptsize
\renewcommand{\arraystretch}{1.2}
\begin{tabular}{llrr}
\hline
                                                                                                       &                                        & \multicolumn{1}{c}{CoQA} & \multicolumn{1}{c}{Synthetic} \\ \hline
\multicolumn{1}{l|}{\multirow{3}{*}{\begin{tabular}[c]{@{}l@{}}Conversational\\ Connectivity\end{tabular}}}     & \multicolumn{1}{l|}{Dependent}           & 68.0\%                   & 66.7\%\,\,                  \\ \cline{2-4} 
\multicolumn{1}{l|}{}                                                                                  & \multicolumn{1}{l|}{Independent} & 27.8\%                    & 28.0\%\,\,                   \\ \cline{2-4} 
\multicolumn{1}{l|}{}                                                                                  & \multicolumn{1}{l|}{Unnatural}             & 4.1\%                    & 5.4\%\,\,                  \\ \hline                                                                          

\multicolumn{1}{l|}{\multirow{2}{*}{\begin{tabular}[c]{@{}l@{}}Question\\ Answerability\end{tabular}}} & \multicolumn{1}{l|}{Answerable}        & 95.7\%                   & 92.6\%\,\,                  \\ \cline{2-4} 
\multicolumn{1}{l|}{}                                                                                  & \multicolumn{1}{l|}{Unanswerable}      & 4.3\%                    & 7.4\%\,\,                  \\ \hline
\multicolumn{1}{l|}{\multirow{3}{*}{\begin{tabular}[c]{@{}l@{}}Answer\\ Correctness\end{tabular}}}     & \multicolumn{1}{l|}{Correct}           & 87.1\%                   & 81.1\%\,\,                  \\ \cline{2-4} 
\multicolumn{1}{l|}{}                                                                                  & \multicolumn{1}{l|}{Partially correct} & 6.5\%                    & 8.4\%\,\,                   \\ \cline{2-4} 
\multicolumn{1}{l|}{}                                                                                  & \multicolumn{1}{l|}{Incorrect}             & 6.4\%                    & 10.5\%\,\,                  \\ \hline
\end{tabular}
\caption{\label{tab:human-evaluation} Human evaluation results of human-annotated data (CoQA) and synthetic data.}
\end{table}

We randomly extracted 100 Q--A pairs with their passages and conversation history from CoQA and the synthetic dataset generated by MultiCQAG. Then, we asked three volunteers to assess 200 examples in terms of the items listed in Table \ref{tab:assessment items}. According to Table \ref{tab:human-evaluation}, there are few grammatical errors or repetitions in the synthetic questions. Additionally, only 3.1\% more unanswerable questions and 4.1\% more incorrect answers were found in synthetic data than in CoQA. From these results, we conclude that MultiCQAG synthesizes data having characteristics similar to human-annotated data.

\subsection{Training Data for Answerability Classification}

We alleviated the problem of the lack of unanswerable samples in CoQA using transfer learning. When the model was trained only with CoQA, it showed a recall of 49.0\% for unanswerable data, as shown in Table \ref{tab:Performance of QAC}. However, when the model trained on QNLI was fine-tuned with CoQA, the recall increased significantly by 27.8\%, although the score for answerable data decreased slightly. The results show that single-turn QA-based QNLI data are helpful in determining the answerability of conversational questions.

\begin{table}[h!]
\centering
\scriptsize
\renewcommand{\arraystretch}{1.2}
\begin{tabular}{lcc}
\hline
Training dataset                 & Answerable-Recall & Unanswerable-Recall \\ \hline
QNLI                             & 74.7            & \textbf{87.4}              \\
CoQA                             & \textbf{99.6}            & 49.0              \\
\textbf{QNLI $\rightarrow$ CoQA} & \underline{98.6}           & \underline{76.8}             \\ \hline
\end{tabular}%
\caption{\label{tab:Performance of QAC} Recall of AC on CoQA development set. The best results are in bold, and the second-best results are underlined.}
\end{table}

\section{Conclusion}

In this paper, we introduce a multi-type data synthesis framework for CQA with individual generation flows for open-ended, closed-ended, and unanswerable questions.
Our framework has a unique two-level classification module to filter invalid Q--A pairs and produce unanswerable questions simultaneously.
By CQA system training and manual evaluations, we proved that the data synthesized with our framework have a quality comparable to that of human-generated CQA data.

\section*{Acknowledgements}
This research was supported by the MSIT (Ministry of Science and ICT), Korea, under the ITRC (Information Technology Research Center) support program (IITP-2022-2020-0-01789) supervised by the IITP (Institute for Information \& Communications Technology Planning \& Evaluation), and also supported by Institute for Information \& communications Technology Planning \& Evaluation (IITP) grantfunded by the Korea government (MSIT) (No. 2021-0-00354, Artificial intelligence technology inferring issues and logically supporting facts from raw text).

\bibliography{anthology,custom}
\bibliographystyle{acl_natbib}

\clearpage

\appendix
\onecolumn

\section{\label{tab:data examples}Examples of Synthetic Data}

\begin{table}[h!]
\centering
\small
\renewcommand{\arraystretch}{1.2}
\begin{tabular}{@{\extracolsep{\fill}}p{16cm}@{\extracolsep{\fill}}}
\hline
\textbf{\textit{Passage}} 
(CNN) -- Colleen LaRose, the Pennsylvania woman indicted for allegedly conspiring to support terrorists and kill a person in a foreign country, attempted to commit suicide in 2005, according to a police report filed at the time. LaRose, who authorities say called herself "Jihad Jane," was depressed about the death of her father, the report from Pennsburg, Pennsylvania, Police Officer Michael Devlin said. LaRose told Devlin she swallowed as many as 10 pills of cyclobenzaprine, a muscle relaxant. The pills were mixed with alcohol. "Colleen was highly intoxicated and having difficulty maintaining her balance," Devlin wrote. I "questioned LaRose about harming herself, at which point she stated she does not want to die." Devlin was dispatched to check on LaRose in response to a 911 call made by LaRose's sister in Texas, who was worried LaRose might try to kill herself. ...\\ \\
\textbf{\textit{Conversation}} \\
Who was indicted for conspiring to support terrorists? Colleen LaRose \\
What state is she from? Pennsylvania \\
When did she attempt suicide? 2005 \\
According to what? a police report filed at the time \\
What did she call herself at the time? Jihad Jane \\
Did she have a boyfriend at the time? unknown \\
What was she depressed about? the death of her father \\
Did she try to kill herself? yes \\
How many pills did she take? as many as 10 pills \\
What was the drug? cyclobenzaprine \\
What was it? a muscle relaxant \\
Did she take it alone? no \\
\hline
\end{tabular}%
\caption*{Example of a passage from CoQA and a conversation generated by MultiCQAG based on the passage.}
\end{table}

\begin{table}[h!]
\centering
\small
\renewcommand{\arraystretch}{1.2}
\begin{tabular}{@{\extracolsep{\fill}}p{16cm}@{\extracolsep{\fill}}}
\hline
\textbf{\textit{Passage}} 
When the love child of the doughnut and the croissant was created by the Dominique Ansel Bakery in New York, {\color{red} fans queued for hours} to sample the tasty hybrid snack.  $\cdot \cdot \cdot$ \\ \\
\textbf{\textit{Conversation history}} \\
Who created the love child of doughnut and croissant? Dominique Ansel Bakery. \\
Where? New York. \\ \\
\textbf{\textit{Question}} Did people queue to try it?   \\
\textbf{\textit{Answer}} Yes.   \\
\hline
\textbf{\textit{Passage}} 
$\cdot \cdot \cdot$ To a friend of more than 20 years, Manssor Arbabsiar was a man who liked to be called "Jack" and didn't seem to have strong views on {\color{red} politics or religion.} To U.S. authorities, the 56-year-old naturalized U.S. citizen is a suspect in an alleged Iranian plot to assassinate Saudi Arabia's ambassador to the United States. $\cdot \cdot \cdot$ \\ \\
\textbf{\textit{Conversation history}} \\
Who is this article about? Manssor Arbabsiar. \\
What did he like to be called? Jack. \\ \\
\textbf{\textit{Question}} Did he have strong opinions on anything?   \\
\textbf{\textit{Answer}} No.   \\
\hline
\end{tabular}%
\caption*{\label{tab:out-of-context} The above table shows examples of closed-ended questions and their answers. The phrases in red are the reference spans extracted from the CAE module.}
\end{table}

\begin{table}[h!]
\centering
\small
\renewcommand{\arraystretch}{1.2}
\begin{tabular}{@{\extracolsep{\fill}}p{16cm}@{\extracolsep{\fill}}}
\hline
\textbf{\textit{Passage}} 
Wiltshire is a county in South West England with an area of . \textbf{It is landlocked and borders the counties of Dorset, Somerset, Hampshire, Gloucestershire, Oxfordshire and Berkshire.} The county town was originally Wilton, after which the county is named, but Wiltshire Council is now based in the county town of Trowbridge. $\cdot \cdot \cdot$ \\
$\cdot \cdot \cdot$ The city of Salisbury is notable for its mediaeval cathedral. {\color{red}Important country houses open to the public include Longleat, near Warminster, and the National Trust's Stourhead, near Mere.} \\  \\
\textbf{\textit{Conversation history}} \\
Is Wiltshire a city? no. \\
What is it notable for? its mediaeval cathedral. \\ \\
\textbf{\textit{Question}} Is it landlocked?   \\
\hline
\textbf{\textit{Passage}} 
Roger Federer and Serena Williams have been named as 2009' s world champions by the International Tennis Federation(ITF) {\color{red}after topping the year-end rankings.} \textbf{Federer, who wins the honour for the fifth time, completed a career Grand Slam at Roland Garros before winning his 15th Grand Slam ride at Wimbledon.} $\cdot \cdot \cdot$ \\ \\
\textbf{\textit{Conversation history}} \\
Who were named 2009 s world champions? Roger Federer and Serena Williams. \\
By who? International Tennis Federation (ITF). \\ \\
\textbf{\textit{Question}} How many times has Federer won this title?   \\
\hline
\end{tabular}%
\caption*{\label{tab:out-of-context} The above table shows examples of \emph{answerable in different context}. The phrased in red are the context sentences that contain the wrong synthetic answers, and the phrases in bold are the correct contexts that contain the actual answers to the questions.}
\end{table}

\begin{table}[h!]
\centering
\small
\renewcommand{\arraystretch}{1.2}
\begin{tabular}{@{\extracolsep{\fill}}p{16cm}@{\extracolsep{\fill}}}
\hline
\textbf{\textit{Passage}} 
$\cdot \cdot \cdot$ Roald Dahl was born in 1916 in Wales, Britain. His father was {\color{red}rich} but he died when Roald was very young. Roald and his mother lived a hard life. $\cdot \cdot \cdot$ \\ \\
\textbf{\textit{Conversation history}} \\
When was Roald Dahl born? 1916. \\
In what country? Britain. \\ \\
\textbf{\textit{Question}} What was his father's occupation?   \\
\hline
\textbf{\textit{Passage}} 
$\cdot \cdot \cdot$ Mr. Clinton and his 13-year-old son Tony are baseball fans. Last October 10th was Tony's birthday, so Mr. Clinton decided to drive him to {\color{red} New York}, for the first game of the World Series . They had no ticket but hoped to buy a pair from others. $\cdot \cdot \cdot$ \\ \\
\textbf{\textit{Conversation history}} \\
What's his name? Tony. \\
What's his age? 13. \\
Is he a baseball fan? Yes. \\ \\
\textbf{\textit{Question}} Where was he from?   \\
\hline
\end{tabular}%
\caption*{\label{tab:out-of-context} The above table shows examples of \emph{unanswerable} question. The phrases in red are the answers synthesized along with the question.}
\end{table}

\newpage

\section{\label{tab:hyperparameters} Training and Data Generation Details}

\begin{table*}[h!]
\centering
\resizebox{\textwidth}{!}{%
\scriptsize
\renewcommand{\arraystretch}{1.2}
\begin{tabular}{l|cccccc}
\hline
Module                                                 & Pretrained model              &              & Epoch & Batch size & Learning rate & Warmup \\ \hline \hline
CAE                           & Bert-large-cased              &              & 2     & 16         & 3e-5          & 0.1    \\ \hline
CQG-AR                    & T5-large                      &              & 3     & 4          & 3e-5          & 0.1      \\ \hline
\multirow{2}{*}{AC} & \multirow{2}{*}{Albert-large} & Pre-training & 10    & 16         & 8e-6          & 0.05   \\
                                                       &                               & Fine-tuning  & 2     & 4          & 1e-6          & 0      \\ \hline
CQA                      & T5-large                      &              & -     & 16         & 3e-5          & 0.1      \\ \hline                                                       
\end{tabular}%
}
\end{table*}
We implemented the modules in the MultiCQAG and CQA systems in Pytorch 1.7 \cite{paszke2019pytorch} and Transformers 4.8.2 \cite{wolf2019huggingface} and used pre-trained language models released from Hugging Face\footnote{https://huggingface.co/} to initialize them. All training and data generation was performed using A100 GPUs. To optimize the models, we used AdamW \cite{loshchilov2018decoupled} with a learning rate scheduler using warm-up steps followed by linear decay. 

In the CQG-AR module, we specify the range of the entire front of the answer span and up to 32 words after it as the answer context. For the AC model, we used a $\tau$ of 0.5, and varied the hyperparameters during pre-training and fine-tuning, as shown in the above table. During data generation, we used a beam search algorithm with a beam size of 4 for the CQG-AR module following \citet{hwang2022conversational}. We also used this decoding strategy during the inference of CQA systems. In addition, we set the ratio of answer types (open-ended:yes:no) to 8:1:1, considering the data distribution of CoQA (Table \ref{tab:Data Distribution}). By generating synthetic data with a distribution of data types that is similar to CoQA, we minimized the impact of differences in these distributions on data quality comparisons.

\end{document}